%% file: main.tex
\documentclass{ieeeaccess}
\usepackage{cite}
\usepackage{amsmath,amssymb,amsfonts}
\usepackage{algorithmic}
\usepackage{graphicx}
\usepackage{multirow}
\usepackage{textcomp}
\usepackage{comment}
\usepackage{hyperref}
\usepackage{cleveref}
\usepackage{makecell}
\usepackage{arydshln} 
\usepackage{xcolor}

\newcommand{\orcid}[1]{%
    \href{https://orcid.org/#1}{\includegraphics[width=10pt]{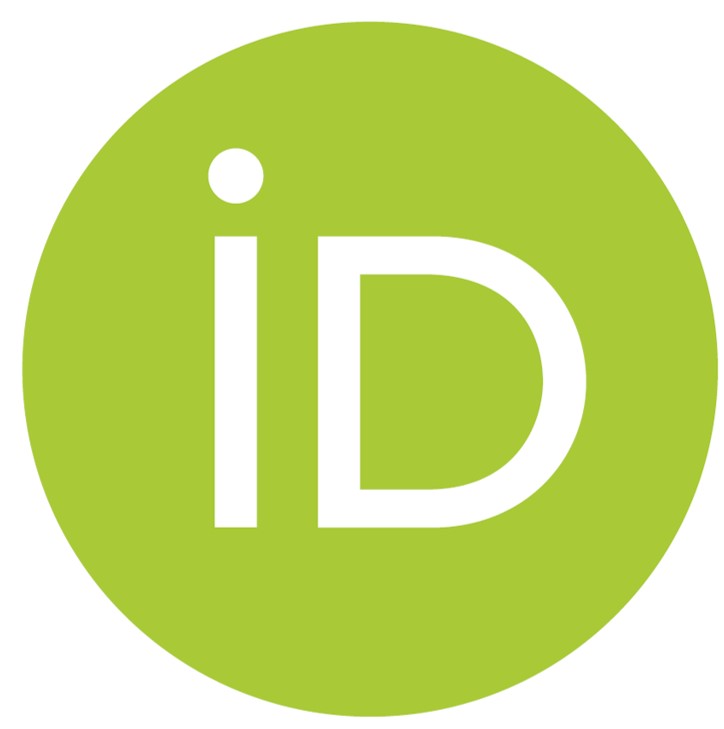}}%
}

\def\BibTeX{{\rm B\kern-.05em{\sc i\kern-.025em b}\kern-.08em
    T\kern-.1667em\lower.7ex\hbox{E}\kern-.125emX}}
\begin{document}

\crefname{figure}{Fig.}{Figs.}
\crefname{table}{Table.}{Tables.}

\title{Toward Efficient Generalization in 3D Human Pose Estimation via a Canonical Domain Approach}

\author{\uppercase{Hoosang Lee}\orcid{0000-0003-0396-8167}\authorrefmark{1} and Jeha Ryu\orcid{0000-0003-4084-5684}\authorrefmark{1} \IEEEmembership{Member, IEEE}}

\address[1]{School of Integrated Technology, Gwangju Institute of Science and Technology (GIST), Gwangju, 61005 Korea Republic (e-mail: hoosang223@gm.gist.ac.kr, ryu@gist.ac.kr)}

\tfootnote{This research was financially supported by the Ministry of Small and Medium-sized Enterprises(SMEs) and Startups(MSS), Korea, under the “Regional Specialized Industry Development Plus Program(R\&D, S3365742)” supervised by the Korea Technology and Information Promotion Agency (TIPA).}

\markboth
{Lee \headeretal: Toward Efficient Generalization in 3D Human Pose Estimation via a Canonical Domain Approach}
{Lee \headeretal: Toward Efficient Generalization in 3D Human Pose Estimation via a Canonical Domain Approach}

\corresp{Corresponding author: Jeha Ryu (e-mail: ryu@gist.ac.kr).}

\input{sec/0_abstract}


\titlepgskip=-21pt

\maketitle

\input{sec/1_introduction}
\input{sec/2_related_work}
\input{sec/3_method}
\input{sec/4_experiment}
\input{sec/5_result_and_discussion}
\input{sec/6_conclusion}
\input{sec/7_acknowledgement}

\bibliographystyle{IEEEtran}
\input{main.bbl}

\EOD

\end{document}

%% file: sec/0_abstract.tex
\begin{abstract}
Recent advancements in deep learning methods have significantly improved the performance of 3D Human Pose Estimation (HPE). However, performance degradation caused by domain gaps between source and target domains remains a major challenge to generalization, necessitating extensive data augmentation and/or fine-tuning for each specific target domain. To address this issue more efficiently, we propose a novel canonical domain approach that maps both the source and target domains into a unified canonical domain, alleviating the need for additional fine-tuning in the target domain. To construct the canonical domain, we introduce a canonicalization process to generate a novel canonical 2D-3D pose mapping that ensures 2D-3D pose consistency and simplifies 2D-3D pose patterns, enabling more efficient training of lifting networks. The canonicalization of both domains is achieved through the following steps: (1) in the source domain, the lifting network is trained within the canonical domain; (2) in the target domain, input 2D poses are canonicalized prior to inference by leveraging the properties of perspective projection and known camera intrinsics. Consequently, the trained network can be directly applied to the target domain without requiring additional fine-tuning. Experiments conducted with various lifting networks and publicly available datasets (e.g., Human3.6M, Fit3D, MPI-INF-3DHP) demonstrate that the proposed method substantially improves generalization capability across datasets while using the same data volume.
\end{abstract}

%% file: sec/1_introduction.tex
\section{Introduction}\label{sec:introduction}
Human pose estimation (HPE) is a computer vision task that identifies and localizes human joints (i.e., keypoints) in images or videos. HPE is an essential area of research and application due to its importance across several domains, including Human-Computer Interaction (HCI), Human-Robot Interaction (HRI), Rehabilitation, Healthcare, Surveillance and Security, Entertainment and Media, Sports Analytics, and Robotics. Recently, deep learning (DL)-based HPE approaches have been actively investigated to improve pose accuracy as well as generalization capability, to handle complex poses and occlusions, or to achieve real-time performance. 

The lack of generalization capability remains one of the major challenges in data-driven approaches involving DL-based HPE. This limitation arises mainly from the domain gap between the source and target data domains, on which the model is trained and tested, respectively. As a result, HPE accuracy deteriorates in cross-dataset evaluations, where the source and target domains originate from different environments \cite{li2020cascaded, gong2021poseaug, huang2022dh, guan2023posegu, li2023cee, peng2024dual, gholami2022adaptpose, chai2023global, liu2023posynda}. To overcome the domain gap problem, research has predominantly focused on two main approaches. The first approach is Domain Generalization (DG) \cite{li2020cascaded, gong2021poseaug, huang2022dh, guan2023posegu, li2023cee, peng2024dual}, which seeks to enhance data diversity through data augmentation techniques, thereby expanding the source domain to cover the target domain. However, addressing diverse environments without prior information of the target domain necessitates generating a substantial volume of diverse data, leading to data inefficiency and demands for larger model capacity. The second approach is \textit{Domain Adaptation} (DA) \cite{gholami2022adaptpose, chai2023global, liu2023posynda}, which leverages available target domain data (i.e., input 2D poses) to align the source domain with the target domain. However, this method requires test-time adaptation, which involves significant time for data augmentation and fine-tuning for each new target domain. Consequently, DG and DA methods often require significant time and effort.

To address the domain gap problem more efficiently, we propose a novel canonical domain approach that transforms both the source and target domains into a unified canonical domain. While this strategy is similar to conventional DA methods, which align the source domain with the target domain, our approach transforms both domains into a unified \textit{canonical domain}, thereby alleviating the need for additional fine-tuning in the target domain.

To construct the canonical domain, we introduce a canonicalization process that generates a canonical 2D-3D pose mapping by rotating the 3D pose around a specified axis to center it on the camera’s principal axis. This design of the canonicalization process offers the following benefits: (1) It narrows the range of 2D and 3D pose distributions while addressing the issue of 2D-3D pose inconsistency, thereby reducing the complexity of the 2D-to-3D lifting problem and facilitating more efficient training of lifting networks on simplified data patterns; (2) it ensures that the same canonical 2D pose can be derived from both ground truth 3D pose and corresponding 2D pose in the normalized image plane, enabling the 2D canonicalization in the target domain where ground truth 3D poses are unavailable.

The canonicalization of both domains is achieved through the following steps: (1) for the source domain, training a lifting network in the canonical domain, and (2) for the target domain, canonicalizing the input 2D pose prior to inference. By training the lifting network on the constructed canonical domain, the source domain is inherently transformed into the canonical domain. To transform the target domain into the canonical domain, the target 2D pose is first mapped to the normalized image plane using known camera parameters, followed by the application of the proposed canonicalization process. With canonicalized target inputs, the lifting network trained on the canonical domain can be applied directly, alleviating the need for further fine-tuning on the target domain.

Using publicly available datasets, including Human3.6M \cite{ionescu2013human3}, Fit3D \cite{Fieraru_2021_CVPR}, and MPI-INF-3DHP \cite{mono-3dhp2017}, we demonstrate that the proposed canonical domain approach significantly enhances pose estimation accuracy with the same data volume, thereby improving data efficiency in cross-dataset evaluations across a variety of lifting networks.

The main contributions can be summarized as follows:
\begin{itemize}
    \item We propose a novel canonical domain approach that maps both the source and target domains into a unified canonical domain, alleviating the need for additional fine-tuning in the target domain. 
    \item We design a canonicalization process to construct the canonical domain, which facilitates more effective training of 2D-3D lifting networks. During inference, this process enables the canonicalization of 2D poses in the target domain without needing ground truth 3D poses.
    \item We provide a mathematical analysis demonstrating how the proposed canonical domain ensures 2D-3D pose consistency and facilitates efficient learning.
    \item We demonstrate the effectiveness of the proposed method in cross-dataset contexts using publicly available datasets and various lifting networks.
\end{itemize}

%% file: sec/2_related_work.tex
\section{Related Work}

\subsection{Deep learning-based Human Pose Estimation} \label{sec:3dhpe}
\noindent \textbf{2D HPE vs 3D HPE} \ DL-based HPE can be broadly classified into two streams based on complexity: \textit{2D HPE} \cite{toshev2014deeppose, Newell2016StackedHN, cao2017realtime, fang2017rmpe, chen2018cascaded, jiang2023rtmpose, rui2023editehrnet}, which involves detecting 2D poses in images, and \textit{3D HPE} \cite{li20153d, li2015maximum, tekin2016structured, pavlakos2017coarse, sun2017compositional, pavlakos2018ordinal, cai2024enetoend, martinez2017simple, pavllo20193d, hossain2018exploiting, fang2018learning, cai2019exploiting, zhao2019semantic, ci2020locally, yu2023gla, zheng20213d, zhang2022mixste, li2022mhformer, zhu2023motionbert, shan2023diffusion, feng2023diffpose, gong2023diffpose, holmquist2023diffpose}, which estimates 3D poses, offering a more detailed and comprehensive view of human posture. 

\noindent \textbf{Frame Input vs Video Input} \ DL-based HPE is also categorized by input type: \textit{frame input} or \textit{video input}. Recent studies \cite{pavllo20193d, hossain2018exploiting, zheng20213d, zhang2022mixste, li2022mhformer, zhu2023motionbert} have utilized video input to exploit temporal relationships between frames for addressing the depth ambiguity problem inherent in frame-based inputs.

\noindent \textbf{Direct Method vs 2D-to-3D Lifting} \ 3D HPE is broadly classified into \textit{Direct method} and \textit{2D-to-3D Lifting}. Direct method \cite{li20153d, li2015maximum, tekin2016structured, pavlakos2017coarse, sun2017compositional, pavlakos2018ordinal, cai2024enetoend} estimates 3D poses directly from images in an end-to-end manner, whereas 2D-to-3D lifting \cite{martinez2017simple, pavllo20193d, hossain2018exploiting, fang2018learning, cai2019exploiting, zhao2019semantic, ci2020locally, yu2023gla, zheng20213d, zhang2022mixste, li2022mhformer, zhu2023motionbert, shan2023diffusion, feng2023diffpose, gong2023diffpose, holmquist2023diffpose} estimates 3D poses using a lifting network that takes extracted 2D poses from images as input.

Starting from \cite{martinez2017simple}, 2D-to-3D lifting has shown significantly better performance than traditional rule-based methods and currently achieves the best 3D HPE performance by utilizing various deep learning architectures, such as MLPs \cite{martinez2017simple}, Temporal Convolutional Networks (TCNs) \cite{pavllo20193d}, Recurrent Neural Networks (RNNs) \cite{hossain2018exploiting, fang2018learning}, Graph Convolutional Networks (GCNs) \cite{cai2019exploiting, zhao2019semantic, ci2020locally, yu2023gla}, Transformers \cite{zheng20213d, zhang2022mixste, li2022mhformer, zhu2023motionbert}, and Diffusion Models \cite{shan2023diffusion, feng2023diffpose, gong2023diffpose, holmquist2023diffpose}. 

In this paper, we focus on the 2D-to-3D lifting approach under single-view, video input, and single-person settings to address the 3D HPE problem.

\input{fig/fig1}

\subsection{Generalization Capability in 3D HPE}
\textbf{Domain Generalization} has primarily focused on creating diverse pose data to improve generalization capabilities without relying on information from the target domain. However, it is essential not only to generate diverse poses but also to ensure their plausiblity. PoseAug \cite{gong2021poseaug} introduced a novel data augmentation framework based on Generative Adversarial Network (GAN) to produce both diverse and plausible pose data. In this framework, a discriminator attempts to differentiate between real and augmented (fake) poses, while an augmentor tries to generate realistic poses to deceive the discriminator. Subsequent studies \cite{huang2022dh, chai2023global, li2023cee, peng2024dual} have also adopted this GAN-based framework, proposing novel methods to further enhance the diversity and plausibility of augmented poses. However, endlessly expanding the source domain distribution without accounting for the target domain is inefficient. Instead, our proposed approach confines both source and target domains to a data-intensive canonical domain, effectively reducing the domain gap and improving data efficiency.

\textbf{Domain Adaptation} \cite{gholami2022adaptpose, chai2023global, liu2023posynda}, on the other hand, aims to enhance the target domain performance using available target 2D poses. This approach generates target domain-like data to finetune the lifting network at each new test environment. Among them, PoseDA \cite{chai2023global} introduced a \textit{Global Position Alignment} (GPA) method to reduce the domain gap caused by global position gap. This method estimates the global translation to align the scale and position of the augmented 2D pose with those of the target 2D pose. Then, the lifting network is then fine-tuned on the augmented data to adapt to the target domain. PoSynDA \cite{liu2023posynda} introduced a diffusion-based data augmentation approach that generates multiple 3D pose hypotheses using a diffusion model conditioned on the target 2D pose and selects the one with the smallest reprojection error as the pseudo 3D ground truth for the target 2D pose. Combined with the GPA strategy, PoSynDA achieves state-of-the-art performance in cross-dataset evaluation.

However, these DA methods require extensive augmented data and fine-tuning for each target domain, resulting in a substantial time cost. In contrast, our proposed method reduces time costs by directly applying the model trained on the canonical domain to the target domain without requiring additional fine-tuning, enabled by the canonicalization of target domain inputs.

\subsection{Canonicalization in 3D HPE}
The domain gap in HPE primarily stems from variations in camera viewpoints, necessitating diverse camera-relative pose data to enhance generalization performance. To overcome this issue, some studies \cite{kundu2022uncertainty, kundu2020kinematic, wang2020predicting, novotny2019c3dpo, li2019adversarial, wei2019view} have applied the concept of canonical space to achieve view-invariant representation of 3D poses, simplifying the learning of data patterns. In these studies, a canonical coordinate frame is defined as one of the following: a body-fixed frame \cite{kundu2022uncertainty, wang2020predicting, wei2019view}, a partially body-fixed frame \cite{kundu2020kinematic}, a global frame \cite{novotny2019c3dpo}, or the camera frame of the source dataset \cite{li2019adversarial}. However, these approaches focus only on the canonicalization of 3D poses, neglecting the canonicalization of the corresponding 2D poses and their consistency, which will be explained in \cref{sec:method_proposed_method}. 

In contrast, our approach canonicalizes both 2D and 3D poses to ensure consistency and simplify 2D-3D pose patterns, facilitating a more efficient learning process.

%% file: fig/fig1.tex
\begin{figure*}[t]
    \centering
    \includegraphics[width=1\linewidth]{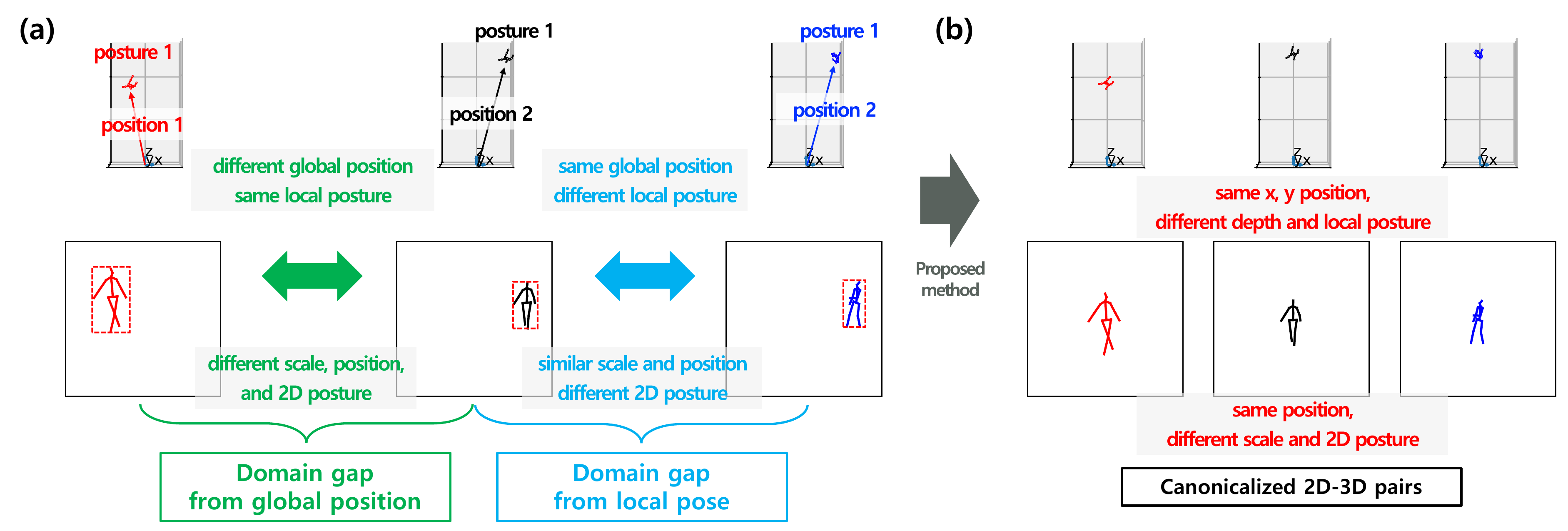}
    \caption{Upper: x-z plane in camera space (top view); Lower: image plane. (a) Global Position vs. Local Pose Domain Gap: Camera-relative 3D poses exhibiting the same posture can result in a domain gap due to global position. This gap is characterized by variations in the scale and position of their corresponding 2D poses. Conversely, a single camera-relative 3D pose with differing postures leads to a domain gap caused by local pose. In this case, the scale and position remain consistent, but the 2D postures vary. (b) Canonicalization of 2D-3D Pose Pairs: Using the proposed method, each 2D-3D pose pair is canonicalized. The root joints of the 3D poses are aligned with the camera’s principal axis, while the 2D poses are repositioned to the image center. This canonicalization preserves variations in both scale and posture.} 
    \label{fig:fig1}
\end{figure*}

%% file: sec/3_method.tex
\section{Method}
\subsection{Preliminary Background} \label{sec:method_prerequisite}
    \subsubsection{Domain Gap in HPE} \label{sec:method_domain_gap}
        In \cite{chai2023global}, the domain gap in HPE is roughly categorized in terms of two aspects: \textit{global position} and \textit{local pose}. \cref{fig:fig1} (a) illustrates both types of domain gap. The global position gap arises from variations in intrinsic and extrinsic camera parameters across datasets. These differences lead to variations in the input 2D pose distribution, particularly in terms of position and scale on the image plane. Here, ‘position’ refers to the x, y coordinates of the root joint, while ‘scale’ refers to the size of the 2D pose. Meanwhile, the local pose varies depending on the diversity of actions, even when position and scale remain the same. Both types of domain gaps have a significant impact on the model’s generalization capability.
        
        Notably, the 2D-to-3D lifting problem involves determining a mapping function from an input 2D pose to an output 3D pose. This task inherently belongs to the \textit{domain shift} problem, where dataset shifts are influenced by both input and output distributions.

    \subsubsection{2D-to-3D Lifting Problem} \label{sec:method_2d_to_3d_lifting}
        In 2D-to-3D lifting problem, a lifting network takes a 2D (skeleton) pose $p \in \mathbb{R}^{J \times 2}$ as input, and outputs a 3D (skeleton) pose $P \in \mathbb{R}^{J \times 3}$, where $J$ denotes the number of joints in the skeleton. Each joint $j$ of a 2D pose has two components, $p^j = [u^j, v^j]^T$, while each joint $j$ of a 3D pose has three components, $P^j = [X^j, Y^j, Z^j]^T$. For simplicity of notation, the superscripts $j$ are omitted in the following equations. We denote the joint $j = r$ as the root joint, which in our case is the \textit{pelvis} joint. In the case of video input, the dimensions of both 2D and 3D skeletons are expanded with the frame dimension $T$ (i.e., $p \in \mathbb{R}^{T \times J \times 2}$ and $P \in \mathbb{R}^{T \times J \times 3}$), facilitating the temporal information of successive poses. Various architectures can be chosen for the lifting network, as mentioned in Section \ref{sec:3dhpe}.

        Ground truth 3D poses are typically captured using motion capture systems (e.g., VICON) and represented in a global frame $G$. A 3D pose in the global frame, $P_G$, can be transformed into a 3D pose in the camera frame $C$, $P_C$, using the camera extrinsic parameters: the rotation matrix $R$ and the translation vector $t$. 
            \begin{equation} \label{eq:world_to_cam}
                P_C = R \times P_G + t = \begin{bmatrix} X_C \\ Y_C \\ Z_C \end{bmatrix}
            \end{equation}
        
        The corresponding 2D pose $p$ can be derived by projecting $P_C$ onto the image plane using the camera intrinsic matrix $K$:
            \begin{equation} \label{eq:original_2d_pose}
                \begin{aligned}
                s \cdot p &= K \times P_C = Z_C \cdot \begin{bmatrix} f_x \frac{X_C}{Z_C} + c_x \\ f_y \frac{Y_C}{Z_C} + c_y \\ 1 \end{bmatrix} = Z_C \cdot \begin{bmatrix}u\\v\\1\end{bmatrix} \\ 
                \therefore \ p &= \begin{bmatrix}u\\v\\1\end{bmatrix} = \begin{bmatrix} f_x \frac{X_C}{Z_C} + c_x \\ f_y \frac{Y_C}{Z_C} + c_y \\ 1 \end{bmatrix} (\text{in homogeneous form}), \\
                \text{where} \ K &= \begin{bmatrix}
                    f_x & 0 & c_x \\
                    0 & f_y & c_y \\
                    0 & 0 & 1
                \end{bmatrix}, \ s: \text{scale factor}.
                \end{aligned}
            \end{equation}
        $f_x$, $f_y$, $c_x$, $c_y$ denote the camera intrinsic parameters, representing the focal lengths in pixels and the image coordinates of the principal axis.
        
        We denote $P_C$ and $p$ as the \textit{original} 3D and 2D pose, respectively, to distinguish them from the proposed \textit{canonical} 3D and 2D pose.
        \input{fig/fig2}
    \subsubsection{Cenventional 2D-3D Mapping and its Canonicalization Strategies} \label{sec:method_conventional_mapping}
        The conventional 2D-3D mapping (\cref{fig:fig3}(b)) used in the 2D-to-3D lifting literature utilizes 2D-3D pose pairs $\{\hat{P}_C, p\}$, where $\hat{P}_C$ represents the root-relative version of the original 3D pose $P_C$. In this conventional mapping, the following canonicalization concepts are employed to enhance lifting performance. We also incorporate these strategies alongside the proposed canonicalization method.

        \noindent \textbf{Root-relative 3D Pose} represents the local position of each joint relative to the root joint, thereby constraining the 3D pose distribution. Thus, it can be considered a form of canonicalization. A root-relative 3D pose in camera frame, $\hat{P}_C$, can be obtained from the $P_C$ as
            \begin{equation} \label{eq:root_relative_3d_pose}
                \hat{P}_C = P_C - P_C^r 
                = \begin{bmatrix} X_C - X_C^r \\ Y_C - Y_C^r \\ Z_C - Z_C^r \end{bmatrix}
                = \begin{bmatrix} \hat{X}_C \\ \hat{Y}_C \\ \hat{Z}_C \end{bmatrix}, 
            \end{equation}
        where $P_C^r$ denotes the root joint of $P_C$.
        
        Generally, the root-relative version of 3D pose is widely used as ground truth during training. This is because the error metric, Mean Per Joint Position Error (MPJPE), calculates the root-relative position error between the ground truth and predictions after aligning the root joint, rather than measuring the absolute joint position error. As a result, it focuses solely on comparing the local posture, disregarding the individual’s absolute position in 3D space.

        \noindent \textbf{Screen Normalization} \footnote{We follow the terminology from \cite{chai2023global}.} is conventionally applied to the 2D pose input to standardize the range of the image plane into the range between -1 and 1, while maintaining the ratio of the 2D pose within the image space. This can also be regarded as a form of canonicalization. The screen normalized 2D pose $p_{norm}$ can be derived as
            \begin{align} \label{eq:screen_norm}
                p_{norm} = (u_{norm}, v_{norm}) = \frac{(u, v) \times 2 - (W, H)}{W},
            \end{align}
        where  $W$ and $H$ represent the width and height of the image, respectively. Note that this process does not change the position, scale, or local posture of the 2D pose in the image plane. 

        For both conventional and canonical mappings, this process is applied to the input 2D pose before it is fed into the network.

\subsection{Proposed Method} \label{sec:method_proposed_method}
    The diverse factors contributing to the domain gap in HPE create an extensive true data distribution that a lifting network must learn to achieve strong generalization capability. Covering this distribution requires a substantial amount of data, as well as significant time and effort.

    To address the domain gap more efficiently, we propose a novel canonical domain approach that transforms both the source and target domains into a unified canonical domain. The canonicalization of both domains is achieved through the following steps: (1) for the source domain, training a lifting network in the canonical domain, and (2) for the target domain, canonicalizing the 2D pose input before inferring the 3D pose using the trained lifting network. 
    
    Before defining the canonical domain, we first identify the issue of 2D-3D pose inconsistency arising from conventional 2D-3D mapping in \cref{sec:method_2d_3d_inconsistency}. Next, \cref{sec:method_canonical_pose_mapping} introduces a canonicalization process that generates consistent and constrained canonical 2D-3D pose mapping, which constructs the canonical domain. Subsequently, \cref{sec:method_effect_of_canonical} presents a mathematical analysis that explains how the proposed canonical domain ensures 2D-3D pose consistency and facilitates efficient learning. Finally, \cref{sec:method_2d_canonicalization} presents a 2D canonicalization process that canonicalizes the target 2D poses without ground truth 3D pose, allowing the trained network on the canonical domain to be fully utilized during inference.
    
    \subsubsection{2D-3D Pose Inconsistency in Conventional 2D-3D Mapping} \label{sec:method_2d_3d_inconsistency}
        While canonicalization strategies in conventional mapping help reduce data distribution size and improve performance to some extent, an additional issue of 2D-3D pose inconsistency arises, where the 3D pose and its corresponding 2D pose exhibit different posture shapes in the x-y plane. We refer to this phenomenon as \textit{2D-3D pose inconsistency}. In \cref{fig:fig2} (a), two distinct positions of the same 3D pose (Position 1 and Position 2) relative to the camera frame are shown. While the root-relative 3D poses remain unchanged in both cases, their 2D projections onto the image plane differ (\cref{fig:fig2} (b)). In Position 1 (blue), the 2D pose retains the original posture shape of the 3D pose in the x-y plane, thereby maintaining 2D-3D pose consistency. In contrast, the 2D pose in Position 2 (red) exhibits a different posture shape (e.g., uncrossed legs), resulting in a lack of 2D-3D pose consistency. 
        
        This inconsistency, which is also difficult for humans to interpret, results from the relative rotation introduced by perspective projection. In perspective projection, all points on an object are projected along lines converging at the camera’s origin. In \cref{fig:fig2} (c), object 1 is centered with the camera principal axis, resulting in its front part (highlighted in blue on object 1) being projected onto the image plane. In contrast, object 2, positioned farther from the principal axis, undergoes a relative rotation with respect to a virtual principal axis passing through its center. Instead of the front part, the left side of object 2 (highlighted in red on object 2) is predominantly projected onto the image plane, creating the appearance of rotation.

        To quantify this inconsistency in 2D pose, consider two 3D poses with the same root-relative 3D pose,  $\hat{P}_C$ , but different positions: one with the pelvis located at Position 1, $[0, 0, Z]^T$ (centered on the principal axis), and the other at Position 2, $[X, Y, Z]^T$  (which has offset from the principal axis in the x and y directions). The 3D poses and their corresponding 2D poses at Position 1 and Position 2 are:
            \begin{equation*}\begin{aligned} 
                P_{C_{pos1}} &= \hat{P}_C + [0, 0, Z]^T = [\hat{X}_C, \hat{Y}_C, Z] 
            \end{aligned}\end{equation*}
            \begin{equation*}\begin{aligned} 
                P_{C_{pos2}} &= \hat{P}_C + [X, Y, Z]^T = [\hat{X}_C + X, \hat{Y}_C + Y, Z] 
            \end{aligned}\end{equation*}
            \begin{equation*}\begin{aligned} 
                p_{pos1} &= \begin{bmatrix} \frac{f_x}{Z} \hat{X}_C + c_x \\ \frac{f_y}{Z} \hat{Y}_C + c_y \end{bmatrix}
            \end{aligned}\end{equation*}
            \begin{equation*}\begin{aligned} 
                p_{pos2} &= 
                \begin{bmatrix} \frac{f_x}{Z} (\hat{X}_C + X) + c_x \\ \frac{f_y}{Z} (\hat{Y}_C + Y) + c_y \end{bmatrix}
            \end{aligned}\end{equation*}
        
        The 2D pose at Position 2, $p_{pos2}$, can be expressed in terms of the 2D pose at Position 1, $p_{C_{pos1}}$:
            \begin{equation*}\begin{aligned}
                 p_{pos2} &= 
                    \begin{bmatrix} \frac{f_x}{Z} (\hat{X}_C + X) + c_x \\ \frac{f_y}{Z} (\hat{Y}_C + Y) + c_y \end{bmatrix}  \\
                    &= \begin{bmatrix} (\frac{f_x}{Z} \hat{X}_C + c_x) + \frac{f_x}{Z} X \\ ( \frac{f_y}{Z} \hat{Y}_C + c_y) + \frac{f_y}{Z} Y \end{bmatrix} 
                    = p_{pos1} + \begin{bmatrix}\frac{f_x}{Z} X \\ \frac{f_y}{Z} Y \end{bmatrix}. 
            \end{aligned}\end{equation*}
        The residual term $\begin{bmatrix}\frac{f_x}{Z} X \\ \frac{f_y}{Z} Y \end{bmatrix}$ between $p_{pos1}$ and $p_{pos2}$ quantifies the effect of relative rotation in the image plane, which contributes to the 2D-3D pose inconsistency. Consequently, this inconsistency arises from the offset to the principal axis in the x and y directions.

        As a result, the model must estimate the same root-relative 3D pose from many different 2D poses while accounting for the residual term, leading to the many-to-one mapping problem. This increases the complexity of the 2D-to-3D lifting problem, further compounded by the inherent challenges of depth ambiguity, thereby necessitating a larger dataset for an effective solution.
    \input{fig/fig3}
    \subsubsection{Canonical 2D-3D Pose Mapping} \label{sec:method_canonical_pose_mapping}
        Similar to constraining 3D poses using the root-relative version, 2D canonicalization can further improve data-efficient learning by constraining the 2D pose distribution. A simple approach to achieve this is translating the original 2D pose to the image center. However, this approach does not consider the issue of 2D-3D pose inconsistency.

        To address this limitation, we propose a novel canonical 2D-3D pose mapping that constrains the 2D-3D data distribution while maintaining 2D-3D pose consistency.
        
        \noindent \textbf{Canonical 3D Pose} Consider an original 3D pose $P_C$ in the camera frame $C$ (blue pose in \cref{fig:fig3} (a)) with its root (pelvis) joint $P_C^r = [X_C^r, Y_C^r, Z_C^r]^T$. A canonical 3D pose $P_{C_{canon}}$ (red pose in \cref{fig:fig3} (a)) can be obtained by centering $P_C$ onto the camera’s principal axis. Specifically, as illustrated in \cref{fig:fig3} (a), the original 3D pose $P_C$ is rotated around an axis defined by the cross product of the principal axis, $\mathbf{v}_{principal} = \mathbf{v}_{z} = [0, 0, 1]^T$, and the pelvis vector, $\mathbf{v}_{pelvis} = P_C^r = [X_C^r, Y_C^r, Z_C^r]^T$. 
        This transformation is achieved using a canonical rotation, $R_{canon}$, which aligns $\mathbf{v}_{pelvis}$ to $\mathbf{v}_{principal}$:
            \begin{align} \label{eq:canoncal_3d_pose}
                P_{C_{canon}} &= \mathbf{R}_{canon} \times P_C 
                = \begin{bmatrix}X_{C_{canon}} \\ Y_{C_{canon}} \\ Z_{C_{canon}}\end{bmatrix}.
            \end{align}
        
        $\mathbf{R}_{canon}$ can be derived using Rodrigues’ rotation formula \cite{rodrigues1840} which defines the rotation matrix $\mathbf{R}$ that aligns vector $\mathbf{a}$ to vector $\mathbf{b}$:
            \begin{align} \label{eq:rodrigues_formula}
                \mathbf{R} &= \mathbf{I} + \sin\theta \mathbf{K} + (1 - \cos\theta)\mathbf{K}^2,
            \end{align}
            \begin{align*}
                \text{where} \ \mathbf{v} &= \mathbf{a} \times \mathbf{b} = [v_x, v_y, v_z]^T \\
                \mathbf{K} &= \begin{bmatrix}
                    0 & -v_z & v_y \\
                    v_z & 0 & -v_x \\
                    -v_y & v_x & 0
                \end{bmatrix} \\
                \sin\theta &= \|\mathbf{a} \times \mathbf{b}\| \\
                \cos\theta &= \mathbf{a} \cdot \mathbf{b}
            \end{align*} 
        In our case, $\mathbf{a}$ and $\mathbf{b}$ correspond to $\mathbf{v}_{pelvis}$ and $\mathbf{v}_{principal}$, respectively. This canonicalization positions the pelvis joint of canonical 3D pose at $[0, 0, \|P_C^r\|]$ along the principal axis, while preserving the original distance from the camera origin to the pelvis joint and local posture of original 3D pose.

        Note that we define canonicalization as a rotation process to enable 2D canonicalization in the target domain, where the original 3D pose is not available, by leveraging the properties of perspective projection, as explained in \cref{sec:method_2d_canonicalization}.
        
        \noindent \textbf{Canonical 2D Pose} Then, by projecting $P_{C_{canon}}$ onto the image plane using the camera intrinsic parameters, the canonical 2D pose $p_{canon}$ can be derived:
            \begin{align} \label{eq:canonical_2d_pose}
                p_{canon}
                &= \begin{bmatrix}
                    u_{canon} \\ v_{canon}
                \end{bmatrix}
                = \begin{bmatrix}
                    f_x \frac{X_{C_{canon}}}{Z_{C_{canon}}} + c_x \\ f_y \frac{Y_{C_{canon}}}{Z_{C_{canon}}} + c_y
                \end{bmatrix} 
            \end{align}
        As illustrated in \cref{fig:fig3} (c), the resulting canonical 2D pose is naturally located on the optical center ($c_x, c_y$) and ensures 2D-3D pose consistency.
        \input{fig/fig4}
        Although the canonicalized 2D pose is derived from the 3D pose centered along the camera’s principal axis, its root joint is not perfectly centered on the image plane due to offsets between the principal point ($c_x, c_y$) and image center ($W/2, H/2$). These offsets depend on the camera’s intrinsic parameters and therefore vary between datasets. To eliminate these offsets and further constrain the data distribution, an additional centering step is applied to the canonical 2D pose after canonicalization.
            \begin{equation} \label{eq:canonical_2d_pose_centered}
                p_{canon} = \begin{bmatrix} u_{canon} \\ v_{canon} \end{bmatrix}
                = \begin{bmatrix}
                    f_x \frac{X_{C_{canon}}}{Z_{C_{canon}}} + W/2 \\ 
                    f_y \frac{Y_{C_{canon}}}{Z_{C_{canon}}} + H/2
                \end{bmatrix} 
            \end{equation}
        This translation preserves the root-relative posture of the 2D pose, thereby maintaining 2D-3D pose consistency. After applying screen normalization as described in \cref{eq:screen_norm}, the constant term ( W/2, H/2 ) is removed, positioning the root joint of the 2D pose at (0, 0) on the image plane. 

        Note that, we do not canonicalize the depth (z-direction) and instead allow it to be inherently learned from the data. As a result, the 2D canonical pose reflects varying scales. This is because the domain gap in the scale of the 2D pose is affected by multiple factors, such as focal length, depth, and variations in human size, which complicate the 2D canonicalization process.

        \noindent \textbf{Canonical 2D-3D Mapping} Finally, we define the canonical 2D-3D pose pairs, $\{\hat{P}_{C_{canon}}, p_{canon}\}$, where $\hat{P}_{C_{canon}}$ denotes the root-relative 3D canonical pose: 
        \begin{equation}\begin{aligned} \label{eq:root_relative_canonical_3d_pose}
                \hat{P}_{C_{canon}} 
                    &= P_{C_{canon}} - P_{C_{canon}}^r \\
                    &= \begin{bmatrix}X_{C_{canon}} \\ Y_{C_{canon}} \\ Z_{C_{canon}} - \|P_C^r\|\end{bmatrix}
                    = \begin{bmatrix}\hat{X}_{C_{canon}} \\ \hat{Y}_{C_{canon}} \\ \hat{Z}_{C_{canon}}^j\end{bmatrix},
            \end{aligned}\end{equation}
        and $P_{C_{canon}}^r = [0, 0, \|P_C^r\|]^T$ denotes the root joint position of the canonical 3D pose. 
        
        \cref{fig:fig3} (b) and (c) show the conventional and proposed canonical 2D-3D mapping. The proposed canonical 2D-3D mapping guarantees 2D-3D pose consistency, whereas the conventional 2D-3D mapping fails to achieve this.     
        In addition, as shown in \cref{fig:fig4}, the canonical pose mapping, compared to conventional mapping, exhibits a similar distribution for root-relative 3D poses while demonstrating a more constrained distribution for 2D poses. This reduction in the 2D pose distribution decreases the complexity of the 2D-3D pose mapping that the model must learn.

        This canonical 2D-3D mapping that ensures 2D-3D consistency constructs the canonical domain and serves as the source domain dataset for training a lifting network. 

    \subsubsection{Effect of Canonical 2D-3D Pose Mapping Compared to Conventional Mapping} \label{sec:method_effect_of_canonical}
        \input{fig/fig5}
        In this section, we compare the proposed 2D-3D pose mapping with the conventional mapping to demonstrate how the proposed method ensures 2D-3D pose consistency and simplifies the learning process.

        First, in the context of conventional mapping, the original 2D pose (\cref{eq:original_2d_pose}) can be rewritten by representing the $X_C$  and $Y_C$ in terms of their root-relative versions, $\hat{X}_C$ and $\hat{Y}_C$, as:
            \begin{equation}\begin{aligned} \label{eq:original_2d_pose_rewrittten}
                p &= \begin{bmatrix} \frac{f_x}{Z_C} X_C + c_x \\ \frac{f_y}{Z_C} Y_C + c_y \end{bmatrix}
                = \begin{bmatrix} \frac{f_x}{Z_C} (\hat{X}_C + X_C^r) + c_x \\ \frac{f_y}{Z_C} (\hat{Y}_C + Y_C^r) + c_y \end{bmatrix} \\
                &= \begin{bmatrix} \frac{f_x}{Z_C} \hat{X}_C + (\frac{f_x}{Z_C}X_C^r + c_x) \\ \frac{f_y}{Z_C} \hat{Y}_C + (\frac{f_y}{Z_C} Y_C^r + c_y) \end{bmatrix},
            \end{aligned}\end{equation}
        where $\hat{X}_C = X_C - X_C^r$ and $\hat{Y}_C = Y_C - Y_C^r$ from \cref{eq:root_relative_3d_pose}. 
        
        From \cref{eq:original_2d_pose_rewrittten,eq:root_relative_3d_pose}, we can derive the conventional 2D-3D mapping:
            \begin{equation} \label{eq:conventional_mapping}
                \text{Conventional}: \begin{bmatrix} \frac{f_x}{Z_C} \hat{X}_C + (\frac{f_x}{Z_C}X_C^r + c_x) \\ \frac{f_y}{Z_C} \hat{Y}_C + (\frac{f_y}{Z_C} Y_C^r + c_y) \end{bmatrix} 
                \mapsto 
                \begin{bmatrix} \hat{X}_C \\ \hat{Y}_C \\ \hat{Z}_C \end{bmatrix}
            \end{equation}
        Second, in the context of canonical mapping, we can rewrite the canonical 2D pose (\cref{eq:canonical_2d_pose_centered}) by representing the $X_{C_{canon}}$  and $Y_{C_{canon}}$ in terms of their root-relative versions, $\hat{X}_{C_{canon}}$ and $\hat{Y}_{C_{canon}}$, as:
            \begin{equation} \label{eq:canonical_2d_pose_rewritten}
                p_{canon}
                = \begin{bmatrix}
                    \frac{f_x}{Z_{C_{canon}}} X_{C_{canon}} + W/2 \\ 
                    \frac{f_y}{Z_{C_{canon}}} Y_{C_{canon}} + H/2
                \end{bmatrix} 
                = \begin{bmatrix}
                    \frac{f_x}{Z_{C_{canon}}} \hat{X}_{C_{canon}} + W/2 \\ 
                    \frac{f_y}{Z_{C_{canon}}} \hat{Y}_{C_{canon}} + H/2
                \end{bmatrix},
            \end{equation}
        where $X_{C_{canon}} = \hat{X}_{C_{canon}}$ and $Y_{C_{canon}} = \hat{Y}_{C_{canon}}$ from \cref{eq:root_relative_canonical_3d_pose}.     
        From \cref{eq:canonical_2d_pose_rewritten,eq:root_relative_canonical_3d_pose}, we can derive the canonical 2D-3D mapping:
            \begin{equation} \label{eq:canonical_mapping}
                \text{Canonical}: \begin{bmatrix} \frac{f_x}{Z_{C_{canon}}} \hat{X}_{C_{canon}} +  W/2\\ \frac{f_y}{Z_{C_{canon}}} \hat{Y}_{C_{canon}} + H/2 \end{bmatrix} \mapsto \begin{bmatrix} \hat{X}_{C_{canon}} \\ \hat{Y}_{C_{canon}} \\ \hat{Z}_{C_{canon}} \end{bmatrix}.
            \end{equation}
        
        By ignoring the known constant terms ($c_x, c_y$) and ($W/2, H/2$) from \cref{eq:conventional_mapping,eq:canonical_mapping}, we derive the final version of both mappings:
        \begin{equation*}\begin{aligned}
            \text{Conventional}: \begin{bmatrix} \frac{f_x}{Z_C} \hat{X}_C + (\frac{f_x}{Z_C}X_C^r) \\ \frac{f_y}{Z_C} \hat{Y}_C + (\frac{f_y}{Z_C} Y_C^r) \end{bmatrix} 
            \mapsto \begin{bmatrix} \hat{X}_C \\ \hat{Y}_C \\ \hat{Z}_C \end{bmatrix} \\
            \text{Canonical}: \begin{bmatrix} \frac{f_x}{Z_{C_{canon}}} \hat{X}_{C_{canon}} \\ \frac{f_y}{Z_{C_{canon}}} \hat{Y}_{C_{canon}} \end{bmatrix} \mapsto \begin{bmatrix} \hat{X}_{C_{canon}} \\ \hat{Y}_{C_{canon}} \\ \hat{Z}_{C_{canon}} \end{bmatrix}.
        \end{aligned}\end{equation*}
        
        In the $x$ and $y$ dimensions, the conventional mapping is expressed in the form  $AX + B \rightarrow X$. Consequently, the conventional mapping requires the model to estimate the offset term $B$, which originates from the offset relative to the principal axis ($X_C^r$, $Y_C^r$), as well as the scale term $A$. This offset induces relative rotation, contributing to the 2D-3D pose inconsistency discussed in \cref{sec:method_2d_3d_inconsistency}. In contrast, the proposed canonical mapping is expressed in the form $AX \rightarrow X$, where only the scale term $A$ needs to be estimated. The absence of the offset term $B$ reflects 2D-3D pose consistency and simplifies the task for the network. This distinction reduces the complexity of the model learning process when using the canonical mapping, as compared to the conventional mapping.

    \subsubsection{2D Canonicalization at Test Time} \label{sec:method_2d_canonicalization}
        Once a lifting network is trained on the canonical domain, it can be directly applied at test time without further fine-tuning, provided that the target domain 2D input is appropriately canonicalized. However, since ground truth 3D poses are not available at test time, canonicalizing the input 2D pose from the 3D pose, as described in \cref{sec:method_canonical_pose_mapping}, is not applicable.
        
        The proposed canonicalization process is designed to address this limitation by leveraging several properties of perspective projection: (1) under perspective projection, each joint has a unique homogeneous coordinate $[\frac{X}{Z}, \frac{Y}{Z}, 1]$, which is obtained by dividing each joint coordinate by its corresponding depth component. This homogeneous coordinate can also be interpreted as the 2D pose projected onto the normalized image plane at a depth of 1 (referred to as the normalized 2D pose); (2) the rotated version of the original 3D joint and its corresponding normalized 2D joint on the normalized image plane also share the same homogeneous coordinates (indicated by the black line on the normalized image plane in \cref{fig:fig5}); and (3) the canonical rotation $R_{canon}$ can be derived from the pelvis vector and the principal axis vector, both of which retain their direction invariance under perspective projection.

        From the first and second properties, given a 3D pose in the camera frame and a rotation matrix $R$, the same 2D pose can be derived from either the rotated 3D pose or the rotated normalized 2D pose by projection with the same intrinsic matrix. This observation is directly applicable to the proposed rotation-based canonicalization process. Therefore, the same canonical 2D pose can be obtained from either the 3D pose or its corresponding normalized 2D pose. From the final property, the same canonical rotation  $R_{canon}$  can also be derived from either representation. As a result, this enables 2D canonicalization in the target domain, where ground truth 3D poses are unavailable, under the mild assumption—as in \cite{liu2023posynda, chai2023global}—that the intrinsic parameters of the target domain are accessible.
    
        The 2D canonicalization process consists of three steps as illustrated in \cref{fig:fig5} (Canonicalization Phase): (1) $K_{target}^{-1}$ normalizes the original target 2D pose ($p^{target}$) into normalized image coordinates, (2) $R_{canon}$ aligns the resulting normalized 2D pose to the principal axis, and (3) $K_{target}$ re-projects it onto the image plane.
            \begin{equation}\begin{aligned} \label{eq:target_canonical_2d}
                s \cdot p_{canon}^{target} &= T_{2D_{canon}} \times p^{target} = s \cdot \begin{bmatrix} u_{canon}^{target} \\ v_{canon}^{target} \\ 1 \end{bmatrix}  \\
                \therefore \ p_{canon}^{target} &= \begin{bmatrix} u_{canon}^{target} \\ v_{canon}^{target} \\ 1 \end{bmatrix} (\text{in homogeneous form}), \\
                \text{where} \ T_{2D_{canon}} &= K_{target} R_{canon} K_{target}^{-1}, 
            \end{aligned}\end{equation}
            \begin{equation*}\begin{aligned}
                K_{target} = \begin{bmatrix}
                        f_x^{tar} & 0 & c_x^{tar} \\
                        0 & f_y^{tar} & c_y^{tar} \\
                        0 & 0 & 1
                    \end{bmatrix}, 
                K_{target}^{-1} = \begin{bmatrix}
                    \frac{1}{f_x^{tar}} & 0 & -\frac{c_x^{tar}}{f_x^{tar}} \\
                    0 & \frac{1}{f_y^{tar}} & -\frac{c_y^{tar}}{f_y^{tar}} \\
                    0 & 0 & 1
                \end{bmatrix}
            \end{aligned}\end{equation*}

        $f_x^{tar}, f_y^{tar}, c_x^{tar}, c_y^{tar}$ represent the intrinsic parameters of the target domain camera.
        In this case, $R_{canon}$ is computed similarly to \cref{eq:canoncal_3d_pose}, with the vector $\mathbf{v}_{pelvis}$ substituted as $[p_x^r, p_y^r, 1]^T$, which is the pelvis point of normalized 2D pose.
        
        After canonicalizing the target domain input, the 2D poses are lifted by the network trained on the canonical domain. The resulting 3D poses in the canonical domain are then back-transformed using the inverse of the canonical rotation matrix, $R_{canon}^{-1}$, to calculate errors against the ground truth 3D poses, which are not in the canonical domain (Inference Phase).

%% file: fig/fig2.tex
\begin{figure*}[t]
    \centering
    \includegraphics[width=1\linewidth]{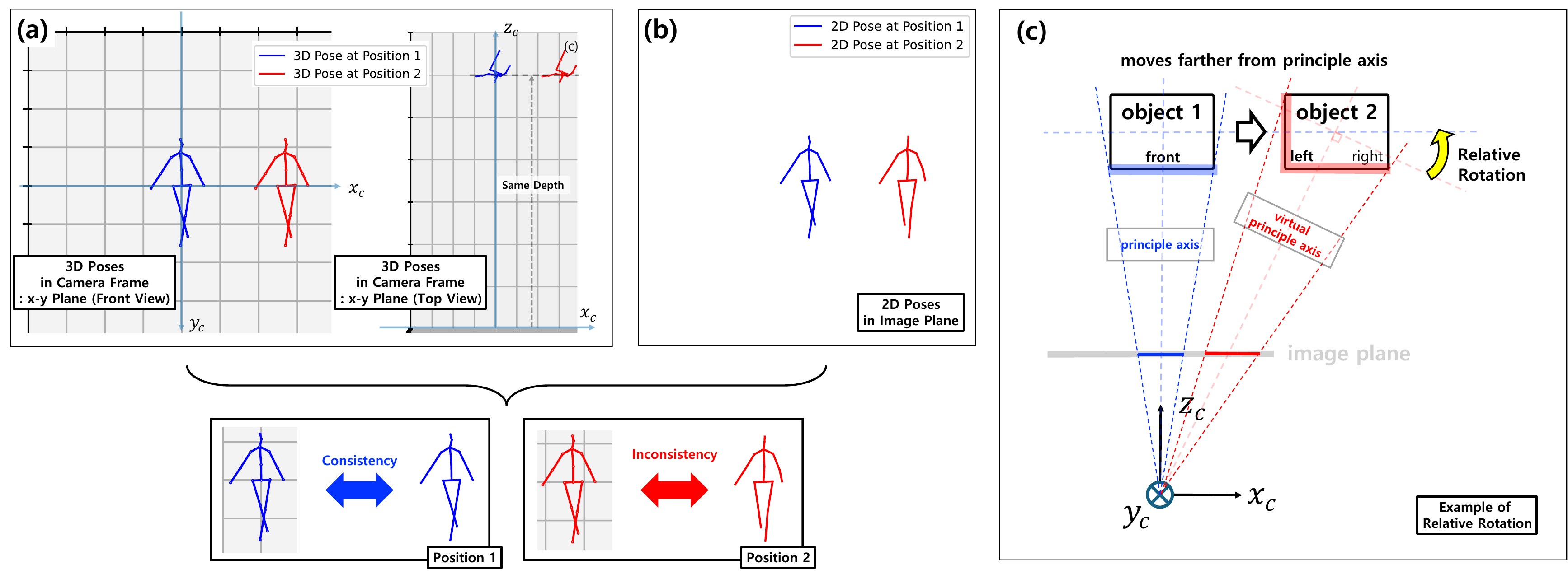}
    \caption{Example of 2D-3D inconsistency: (a) Two 3D poses with identical postures located at different positions (Position 1 and Position 2) in camera space. (b) Corresponding 2D poses projected onto the image plane, illustrating differences in posture shape. (c) An example of relative rotation caused by perspective projection.}
    \label{fig:fig2}
\end{figure*}

%% file: fig/fig3.tex
\begin{figure*}[t]
    \centering
    \includegraphics[width=1\linewidth]{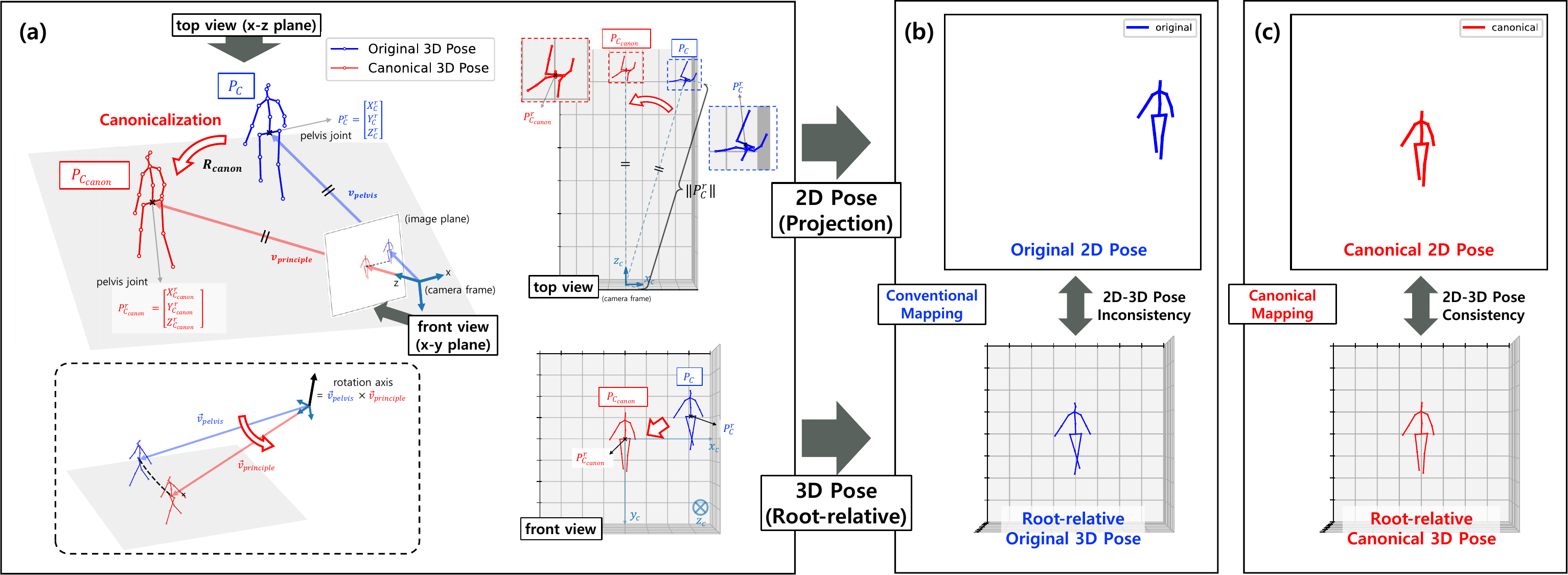}
    \caption{Comparison between conventional and canonical 2D-3D mapping. (a) Proposed Canonicalization Process for 3D Pose, (b) Conventional 2D-3D Mapping, (c) Proposed Canonical 2D-3D Pose Mapping.}
    \label{fig:fig3}
\end{figure*}

%% file: fig/fig4.tex
\begin{figure*}[t]
    \centering
    \includegraphics[width=1\linewidth]{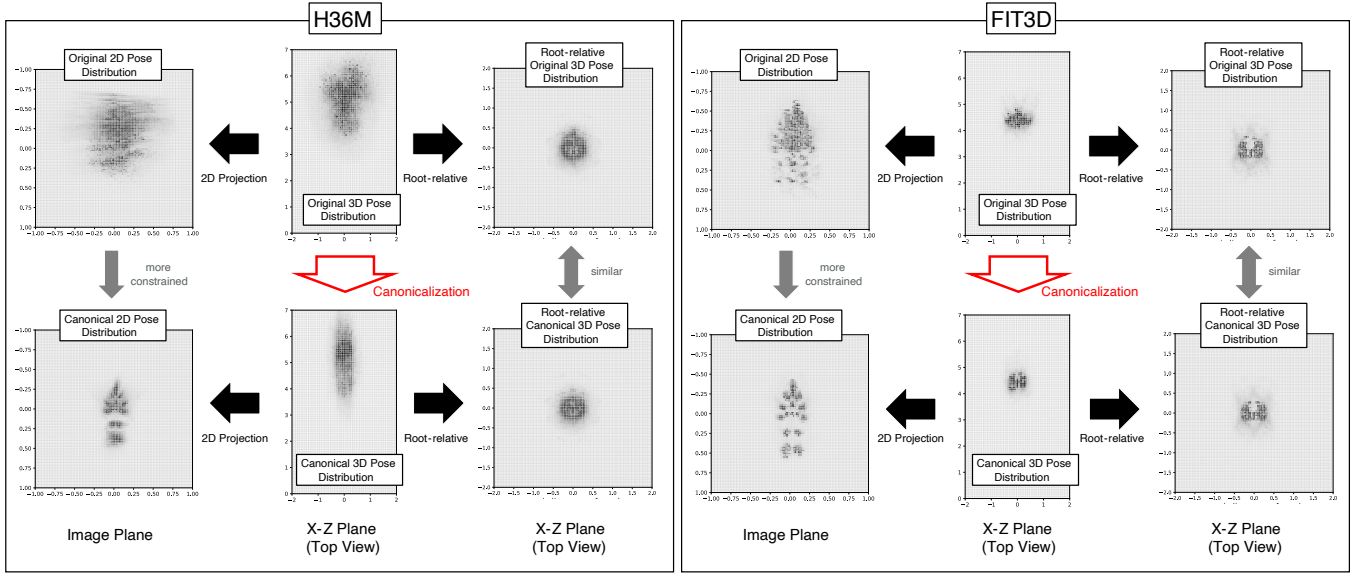}
    \caption{Comparison between Original and Canonical 2D-3D Data Distribution. The (root-relative) 3D pose distributions represent the overall 3D joint positions of the entire dataset in the x-y plane (top view) of the camera frame. The 2D pose distributions depict the overall 2D joint positions of the entire dataset in the image plane.}
    \label{fig:fig4}
\end{figure*}

%% file: fig/fig5.tex
\begin{figure*}[t]
    \centering
    \includegraphics[width=0.8\linewidth]{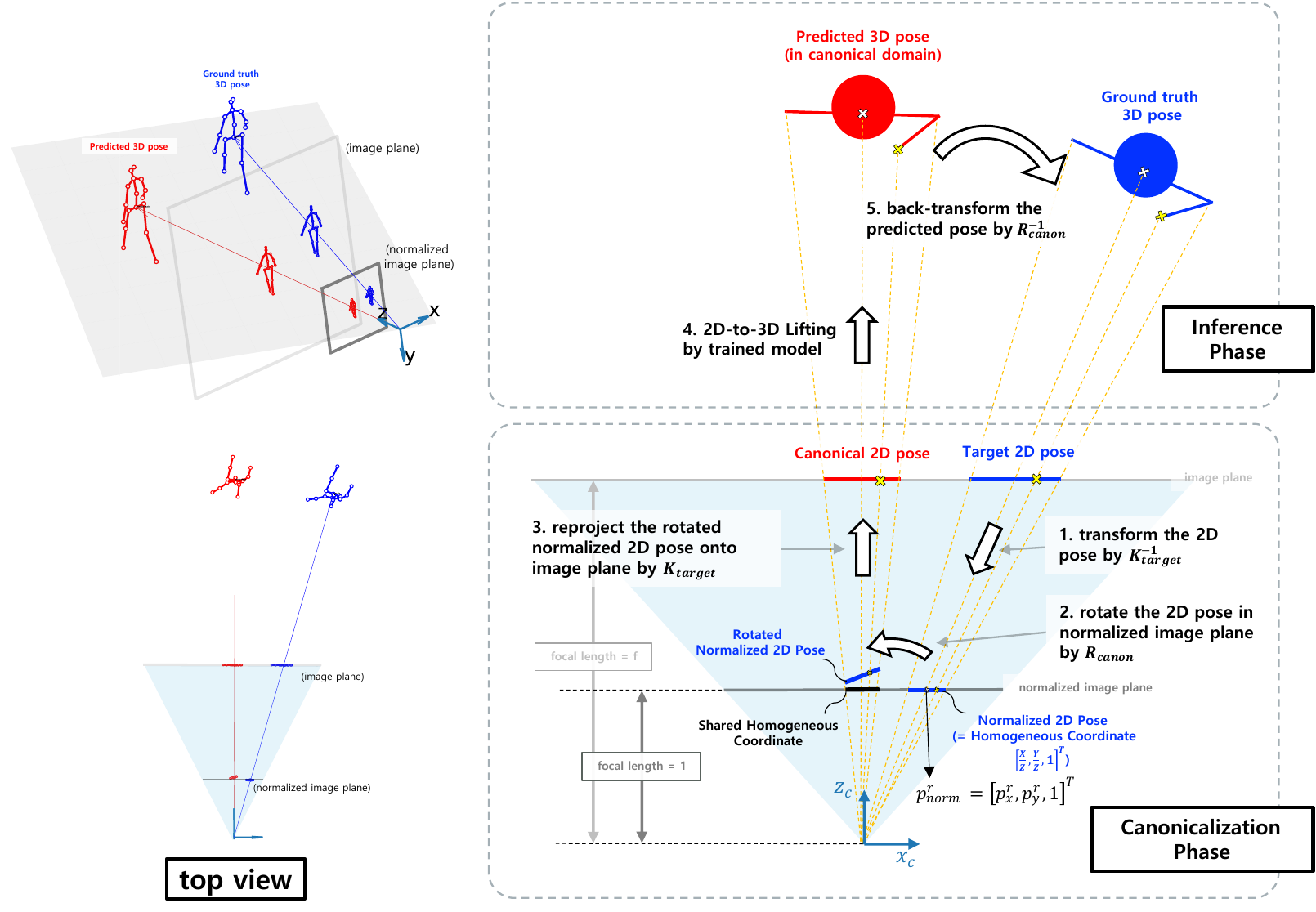}
    \caption{2D Canonicalization and Inference Process: (1) Target 2D poses are transformed into the normalized image plane by $K_{target}^{-1}$; (2) resulting normalized target poses are rotated by $R_{canon}$ that aligns the pelvis vector $[p_x, p_y, 1]$ to principal axis; and then (3) reprojected to image plane by $K_{target}$; (4) the lifting network predicts the 3D poses from the canonical 2D poses; and (5) the predicted 3D poses are back-transformed by $R_{canon}^{-1}$ for comparison with the ground truth.}
    \label{fig:fig5}
\end{figure*}

%% file: sec/4_experiment.tex
\section{Experiment}
\subsection{Dataset}
    To evaluate the effectiveness of the proposed method on HPE performance, we select three datasets: Human3.6M \cite{ionescu2013human3}, Fit3D \cite{Fieraru_2021_CVPR}, and MPI-INF-3DHP \cite{mono-3dhp2017}, providing a comprehensive dataset for training and evaluating lifting networks.    
    
    \noindent \textbf{Human3.6M (H36M)} dataset \cite{ionescu2013human3} is one of the largest publicly available datasets for human pose estimation. It includes approximately 3.6 million 3D human poses with corresponding images. The dataset features 11 professional actors performing 15 diverse daily activities. Each frame includes a 3D skeleton pose of 32 body joints, captured using a motion capture system. The data was recorded in a controlled indoor environment with four calibrated cameras, providing synchronized multi-view video frames. The H36M training set comprises data from 5 subjects (S1, S5, S6, S7, S8), while the test set includes data from 2 subjects (S9, S11).
    
    \noindent \textbf{Fit3D} dataset \cite{Fieraru_2021_CVPR} is specifically designed to provide human movement data for fitness training. It includes data from 13 subjects, comprising one licensed fitness instructor and trainees with varying skill levels. Each frame includes a 3D skeleton pose of 25 body joints. The subjects perform 47 exercises, categorized into four types: warm-ups, barbell exercises, dumbbell exercises, and equipment-free exercises. Data collection is carried out using four synchronized cameras and motion capture equipment, resulting in comprehensive 2D/3D skeleton poses and video recordings. Additionally, the dataset includes manually labeled repetition counts, enhancing its utility for fitness-related applications. For our experiments, we utilize data from 8 subjects (subjects 3, 4, 5, 7, 8, 9, 10, and 11) that provide ground truth 3D poses.
    \input{table/hyperparameters}
    
    \noindent \textbf{MPI-INF-3DHP (3DHP)} dataset \cite{mono-3dhp2017} includes a total of 1.3 million frames of annotated data from 14 camera views with 8 subjects performing 8 activities. Each frame includes a 3D skeleton pose of 17 body joints, captured using a markerless motion capture system. The dataset offers a mix of controlled indoor scenes and challenging outdoor scenes, providing diverse lighting and background conditions. This diversity makes MPI-INF-3DHP particularly useful for cross-dataset testing to assess the generalization performance of 3D HPE model. In this study, we used the 3DHP test set (TS1–TS6) with universal skeleton-based 3D ground truth, as described in \cite{mono-3dhp2017}, enabling a fair comparison with previous DG and DA methods.
    
\subsection{Baseline 2D-to-3D Lifting Method}
\input{table/cross_scenario}
    MotionBERT \cite{zhu2023motionbert} is a transformer-based architecture developed for various human pose-related tasks, including pose estimation, mesh recovery, and action recognition. The authors proposed training a motion encoder on heterogeneous data resources to develop a versatile human motion representation within a unified framework. Inspired by the success of language models trained to recover masked inputs, they hypothesized that reconstructing 3D poses from 2D poses is an effective pretext task for learning robust motion representations. The Dual-stream Spatio-temporal Transformer (DSTformer) is introduced as a motion encoder architecture to capture long-range relationships among sequential human poses. The encoder is trained to reconstruct 3D poses from corrupted 2D poses using various 2D and 3D human pose datasets, including Human3.6M \cite{ionescu2013human3}, AMASS \cite{mahmood2019amass}, PoseTrack \cite{andriluka2018posetrack}, and InstaVariety \cite{kanazawa2019learning}. Subsequently, it is fine-tuned with additional layers for downstream tasks. For 2D datasets without 3D pose ground truth, a weighted 2D re-projection loss is employed. In this work, we adopt the DSTformer architecture as the baseline lifting network to investigate the effect of the proposed method.
    
\subsection{Implementation details} \label{implementation_details}
    We trained the DSTformer network from scratch with hyperparameters in \cref{tab:hyperparameters}. For a fair comparison, all models were trained with the same hyperparameters. The model takes 2D pose sequences $x \in \mathbb{R}^{T \times J \times 2}$ as input and outputs motion features $F \in \mathbb{R}^{T \times J \times C_f}$. The extracted features are then converted into the final estimated 3D pose $X \in \mathbb{R}^{T \times J \times 3}$ by fully-connected layers and \textit{tanh} activation function. We used AdamW, a variant of the Adam optimizer \cite{kingma2014adam} with weight decay, implemented in PyTorch. All training and inference were conducted on an Ubuntu 20.04 system with a Ryzen 5950X CPU and 2 NVIDIA RTX 4090 GPUs. Input data is segmented into 243-frame sequences to match the model’s input size, with an 81-frame stride for training and a 243-frame stride for testing.
    
\subsection{Experiment Details}
    \noindent \textbf{Experiment Type} Each experiment can be categorized into the following two types based on the datasets used for the source and target domains:
    \begin{itemize}
        \item \textbf{Cross-scenario Evaluation}: The source and target domains are derived from the same dataset; therefore, both domains consist of data captured using the same camera setup in the same environment. However, the composition of subjects from whom the data was collected differs between the two domains.
        \item \textbf{Cross-dataset Evaluation}: The source and target domains are derived from different datasets. Consequently, both domains consist of data captured with different camera setups and subjects, making this setup ideal for assessing the model’s generalization capabilities.
    \end{itemize}
    \input{table/cross_dataset}
    \noindent \textbf{Dataset Details} H36M and Fit3D are used for both Cross-scenario and Cross-dataset evaluation, while 3DHP is used for Cross-dataset evaluation. Following the conventional setting for the H36M experiment, we train a model on the H36M training set (S1, 5, 6, 7, 8) and evaluate it on the H36M test set (S9, 11). To investigate the impact of varying amounts of training data, we adopt the partial H36M experimental setting from previous works. In this setting, only one subject (S1) is used for training, and the remaining subjects from the H36M training set (S5, 6, 7, 8) are used for testing. 
    Similarly, for Fit3D, we construct two experimental settings: a 5-subject training set (S3, S5, S8, S9, S11 for training and S4, S7, S10 for testing) and a 1-subject training set (S3 for training and S4, S5, S7, S8, S9, S10, S11 for testing).

    \noindent \textbf{Methods } In all experiments, we compare the following methods, which utilize DSTformer as a common lifting network and differ in the type of 2D-3D pose mapping of the dataset:
    \begin{itemize}
        \item \textbf{Conventional} (baseline): Use conventional 2D-3D pose mapping (\cref{fig:fig3} (b)) = (screen normalized) original 2D pose + root-relative original 3D pose.
        \item \textbf{Canonical} (proposed method): Use canonical 2D-3D pose mapping (\cref{fig:fig3} (c)) = (screen normalized) canonical 2D pose + root-relative canonical 3D pose.
    \end{itemize}
    
    \noindent \textbf{Metrics} For performance metric, Mean Per Joint Position Error (MPJPE) and Procrustes Aligned MPJPE (P-MPJPE) are used. MPJPE calculates the average error between the estimated and the ground truth positions of all joints in the test set. To calculate the joint position error, the root (pelvis) joints of the estimated and ground truth poses are aligned, and the root-relative error is then measured:
    \begin{equation}
        MPJPE = \frac{1}{T} \frac{1}{J} \sum_{t=1}^T \sum_{j=1}^J \parallel \hat{P}_C^{j, t} - \hat{P}_{C_{gt}}^{j, t}  \parallel_2,
    \end{equation}
    where $\hat{P}_C^{j, t}$ and $\hat{P}_{C_{gt}}^{j, t}$ denote the root-relative position of $j$-th joint at frame $t$ and its ground truth, respectively.P-MPJPE is computed similarly to MPJPE, but the orientation and scale of the predicted pose are aligned with the ground truth before error calculation.
    
    For DSTformer, the best performance is selected from models trained using five different random seeds (0, 1, 2, 3, and 4). For the additional lifting networks, the results are obtained from the model trained with the same seed 0.    

%% file: table/hyperparameters.tex
\begin{table}[]
    \centering
    \caption{Hyperparameters for DSTformer Training.}
    \label{tab:hyperparameters}
    \scriptsize
    \begin{tabular}{|l|l|}
        \hline 
        \textbf{Hyperparameter}  & \textbf{Value}  \\ \hline
        epoch           & 60     \\ \hline
        batch size      & 16     \\ \hline
        learning rate   & 0.0002 \\ \hline
        weight decay    & 0.01   \\ \hline
        lr decay        & 0.99   \\ \hline
        frame length ($T$)    & 243    \\ \hline
        feature dimension ($C_f$) & 512 \\ \hline
        number of attention head  & 8      \\ \hline
        number of attention depth & 5      \\ \hline
    \end{tabular}%
\end{table}

%% file: table/cross_scenario.tex
\begin{table*}[]
\centering
\caption{Experiment Results for Cross-scenario Test}
\begin{tabular}{|c|c|c|c|c|c|c|}
\hline
\multirow{2}{*}{Train set} &
  \multirow{2}{*}{Test set} &
  \multirow{2}{*}{Method} &
  \multirow{2}{*}{\begin{tabular}[c]{@{}c@{}}MPJPE \\ (mm) $\downarrow$\end{tabular}} &
  \multirow{2}{*}{\begin{tabular}[c]{@{}c@{}}Error \\ reduction rate $\uparrow$\end{tabular}} &
  \multirow{2}{*}{\begin{tabular}[c]{@{}c@{}}P-MPJPE \\ (mm) $\downarrow$\end{tabular}} &
  \multirow{2}{*}{\begin{tabular}[c]{@{}c@{}}Error \\ reduction rate $\uparrow$\end{tabular}} \\
 &
   &
   &
   &
   &
   &
   \\ \hline
\multirow{2}{*}{\begin{tabular}[c]{@{}c@{}}H36M\\ S1\end{tabular}} &
  \multirow{2}{*}{\begin{tabular}[c]{@{}c@{}}H36M\\ S5, 6, 7, 8\end{tabular}} &
  Conventional &
  42.08 &
  - &
  29.61 &
  - \\ 
 &
   &
  Canonical (Ours) &
  \textbf{40.32} &
  \textbf{4.18\%} &
  \textbf{27.4} &
  \textbf{7.46\%} \\ \hline
\multirow{2}{*}{\begin{tabular}[c]{@{}c@{}}H36M\\ S1, 5, 6, 7, 8\end{tabular}} &
  \multirow{2}{*}{\begin{tabular}[c]{@{}c@{}}H36M\\ S9, 11\end{tabular}} &
  Conventional &
  \textbf{20.44} &
  \textbf{-} &
  \textbf{15.92} &
  \textbf{-} \\ 
 &
   &
  Canonical (Ours) &
  20.85 &
  -2.01\% &
  16.18 &
  -1.63\% \\ \hline
\multirow{2}{*}{\begin{tabular}[c]{@{}c@{}}FIT3D\\ S3\end{tabular}} &
  \multirow{2}{*}{\begin{tabular}[c]{@{}c@{}}FIT3D\\ EXCEPT S3\end{tabular}} &
  Conventional &
  \textbf{39.77} &
  \textbf{-} &
  25.83 &
  - \\ 
 &
   &
  Canonical (Ours) &
  40.32 &
  -1.38\% &
  \textbf{24.42} &
  \textbf{5.46\%} \\ \hline
\multirow{2}{*}{\begin{tabular}[c]{@{}c@{}}FIT3D\\ S3, 5, 8, 9, 11\end{tabular}} &
  \multirow{2}{*}{\begin{tabular}[c]{@{}c@{}}FIT3D\\ S4, 7, 10\end{tabular}} &
  Conventional &
  31.51 &
  - &
  20.54 &
  - \\ 
 &
   &
  Canonical (Ours) &
  \textbf{31.04} &
  \textbf{1.49\%} &
  \textbf{19.49} &
  \textbf{5.11\%} \\ \hline
\end{tabular}
\label{tab:cross_scenario}
\end{table*}

%% file: table/cross_dataset.tex
\begin{table*}[]
\centering
\caption{Experiment Results for Cross-dataset.}
\begin{tabular}{|c|c|c|c|c|c|c|}
\hline
\multirow{2}{*}{Train set} &
  \multirow{2}{*}{Test set} &
  \multirow{2}{*}{Model} &
  \multirow{2}{*}{\begin{tabular}[c]{@{}c@{}}MPJPE \\ (mm) $\downarrow$\end{tabular}} &
  \multirow{2}{*}{\begin{tabular}[c]{@{}c@{}}Error \\ reduction rate $\uparrow$\end{tabular}} &
  \multirow{2}{*}{\begin{tabular}[c]{@{}c@{}}P-MPJPE \\ (mm) $\downarrow$\end{tabular}} &
  \multirow{2}{*}{\begin{tabular}[c]{@{}c@{}}Error \\ reduction rate $\uparrow$\end{tabular}} \\
 &
   &
   &
   &
   &
   &
   \\ \hline
\multirow{4}{*}{\begin{tabular}[c]{@{}c@{}}H36M\\ S1\end{tabular}} &
  \multirow{2}{*}{\begin{tabular}[c]{@{}c@{}}FIT3D\\ ALL\end{tabular}} &
  Conventional &
  158.49 &
  - &
  94.57 &
  - \\ 
 &
   &
  Canonical (Ours) &
  \textbf{110.21} &
  \textbf{30.46\%} &
  \textbf{71.2} &
  \textbf{24.71\%} \\ \cline{2-7} 
 &
  \multirow{2}{*}{\begin{tabular}[c]{@{}c@{}}3DHP\\ TEST SET\end{tabular}} &
  Conventional &
  87.75 &
  - &
  63.85 &
  - \\ 
 &
   &
  Canonical (Ours) &
  \textbf{70.46} &
  \textbf{19.70\%} &
  \textbf{57.53} &
  \textbf{9.90\%} \\ \hline
\multirow{4}{*}{\begin{tabular}[c]{@{}c@{}}H36M\\ S1, 5, 6, 7, 8\end{tabular}} &
  \multirow{2}{*}{\begin{tabular}[c]{@{}c@{}}FIT3D\\ ALL\end{tabular}} &
  Conventional &
  159.14 &
  - &
  101.75 &
  - \\ 
 &
   &
  Canonical (Ours) &
  \textbf{87.64} &
  \textbf{44.93\%} &
  \textbf{70.87} &
  \textbf{30.35\%} \\ \cline{2-7} 
 &
  \multirow{2}{*}{\begin{tabular}[c]{@{}c@{}}3DHP\\ TEST SET\end{tabular}} &
  Conventional &
  64.87 &
  - &
  50.60 &
  - \\ 
 &
   &
  Canonical (Ours) &
  \textbf{59.53} &
  \textbf{8.23\%} &
  \textbf{47.79} &
  \textbf{5.55\%} \\ \hline
\multirow{4}{*}{\begin{tabular}[c]{@{}c@{}}FIT3D\\ S3\end{tabular}} &
  \multirow{2}{*}{\begin{tabular}[c]{@{}c@{}}H36M\\ ALL\end{tabular}} &
  Conventional &
  182.47 &
  - &
  118.91 &
  - \\ 
 &
   &
  Canonical (Ours) &
  \textbf{121.1} &
  \textbf{33.63\%} &
  \textbf{93.73} &
  \textbf{21.18\%} \\ \cline{2-7} 
 &
  \multirow{2}{*}{\begin{tabular}[c]{@{}c@{}}3DHP\\ TEST SET\end{tabular}} &
  Conventional &
  191.73 &
  - &
  132.11 &
  - \\ 
 &
   &
  Canonical (Ours) &
  \textbf{143.85} &
  \textbf{24.97\%} &
  \textbf{106.22} &
  \textbf{19.60\%} \\ \hline
\multirow{4}{*}{\begin{tabular}[c]{@{}c@{}}FIT3D\\ S3, 5, 8, 9, 11\end{tabular}} &
  \multirow{2}{*}{\begin{tabular}[c]{@{}c@{}}H36M\\ ALL\end{tabular}} &
  Conventional &
  177.09 &
  - &
  118.85 &
  - \\ 
 &
   &
  Canonical (Ours) &
  \textbf{113.96} &
  \textbf{35.65\%} &
  \textbf{85.8} &
  \textbf{27.81\%} \\ \cline{2-7} 
 &
  \multirow{2}{*}{\begin{tabular}[c]{@{}c@{}}3DHP\\ TEST SET\end{tabular}} &
  Conventional &
  189.23 &
  - &
  128.84 &
  - \\ 
 &
   &
  Canonical (Ours) &
  \textbf{135.56} &
  \textbf{28.36\%} &
  \textbf{99.63} &
  \textbf{22.67\%} \\ \hline
\end{tabular}
\label{tab:cross_dataset}
\end{table*}

%% file: sec/5_result_and_discussion.tex
\section{Result and Discussion} \label{sec:result_and_discussion}
    \subsection{Cross-scenario Evaluation} \label{sec:cross_scenario_evaluation}
        \input{fig/fig6}
        \input{table/ablation_study}
        \cref{tab:cross_scenario} presents the experimental results for the cross-scenario evaluation. The proposed method exhibits comparable performance to the baseline, with slight variations across scenarios. This outcome is likely attributable to the nature of cross-scenario tests, where the source and target domains share similar characteristics (e.g., camera settings and action types), resulting in a relatively small domain gap. Therefore, the effectiveness of transforming the source and target domains into the canonical domain is relatively limited.
        
        More specifically, when training on H36M, the proposed method demonstrates better improvement in MPJPE and P-MPJPE when trained on a single subject compared to five subjects. This suggests that the proposed method improves data efficiency by facilitating more effective learning from the same data volume. However, when training on Fit3D, this phenomenon is not observed. This is because the Fit3D dataset is already distributed in positions that are nearly identical, with only small offsets relative to the principal axis, as illustrated in \cref{fig:fig6} (a). This property of Fit3D can be interpreted as being positionally canonicalized and maintaining 2D-3D pose consistency. As a result, the impact of the proposed method is diminished in this scenario.

    \subsection{Cross-dataset Evaluation} \label{sec:cross_dataset_evaluation}
        To evaluate the cross-dataset generalization capability that we mainly focused on, we tested four trained models described in \cref{sec:cross_scenario_evaluation} using test sets from target domains that differ from the training set (i.e., H36M → Fit3D, 3DHP; Fit3D → H36M, 3DHP).
          
        In \cref{tab:cross_dataset}, unlike the cross-scenario results, the proposed canonicalization method demonstrates highly significant performance improvements over the baseline in terms of generalization capability. This result underscores the effectiveness of introducing a canonical domain in mitigating the domain gap and improving cross-domain adaptability.
            
        Note that when training with Fit3D, their overall absolute error level is higher compared to training with H36M. These results can be attributed to the characteristics of the Fit3D dataset, which lacks diversity in camera-relative poses (position and orientation) compared to H36M and 3DHP, as illustrated in \cref{fig:fig6}. This approach fails to provide diverse camera-relative poses, leading to overfitting, which is critical for constructing a generalizable canonical domain. This observation underscores the importance of future work on developing data-efficient augmentation techniques aimed at mitigating overfitting by redistributing the source dataset, rather than simply increasing its volume with augmented data.

    \subsection{Ablation Study} \label{sec:ablation_study}
        To verify the effects of 2D canonicalization and 2D-3D pose inconsistency, we further compared the conventional and proposed methods with a simple 2D canonicalization approach, where the conventional 2D-3D mapping is used with the input 2D pose centered on the image plane. We refer to this method as \textit{Conventional + Input Centering (IC)}. This approach constrains the 2D pose distribution similarly to the proposed method but does not address the issue of 2D-3D pose inconsistency.

        \cref{tab:ablation_study} shows that \textit{Conventional+IC} exhibits varying performance in MPJPE errors depending on the test sets. Specifically, \textit{Conventional+IC} achieves significant MPJPE improvements when tested on the FIT3D dataset. This behavior is due to the characteristics of Fit3D, where 2D-3D pose consistency is nearly ensured, as discussed in \cref{sec:cross_scenario_evaluation}. This result emphasizes the inherent benefits of 2D canonicalization itself. However, \textit{Conventional+IC} performs poorly on the 3DHP test set, which contains poses that are more widely scattered from the principal axis and therefore lack 2D-3D pose consistency. Conversely, the proposed method addresses this limitation by enforcing 2D-3D pose consistency during both training and inference, leading to improved MPJPE on the 3DHP test set. This result highlights the importance of 2D-3D pose consistency and underscores its critical role in achieving robust performance across diverse datasets. 
        
        On the other hand, \textit{Conventional+IC} performs well in terms of P-MPJPE error, occasionally achieving slightly better results than the proposed method in certain case (H36M S1,5,6,7,8 → FIT3D ALL). Since P-MPJPE aligns the orientation of the predicted and ground truth 3D poses, it mitigates the positional error caused by 2D-3D pose inconsistency. Consequently, this improvement can be attributed to the effect of 2D canonicalization itself.
    
    \subsection{Evaluation on Additional Lifting Networks} \label{sec:additional_network_evaluation}
    \input{table/additional_network}
        To verify the effectiveness of the proposed method across various lifting networks, we further evaluated it on additional architectures, including VideoPose \cite{pavllo20193d}, SimpleBaseline (SB) \cite{martinez2017simple}, SemGCN \cite{zhao2019semantic}, and ST-GCN \cite{cai2019exploiting} \footnote{We utilized the official code of PoseAug \cite{gong2021poseaug} for model implementation, training, and test.}. The results, presented in \cref{tab:additional_network}, exhibit a trend consistent with the previous cross-scenario and cross-dataset evaluations: our method is more effective in cross-dataset evaluation (3DHP test set) than in cross-scenario evaluation (H36M).

        Notably, in the 3DHP cross-dataset evaluation, the networks trained using our method achieved lower MPJPE errors compared to those trained with the PoseAug method, except for ST-GCN (where the performance is still comparable). This demonstrates that our method exhibits strong generalization capability, comparable to the domain generalization method, without the need for additional data augmentation or extended training. These results highlight the effectiveness of our approach in reducing computational and time resources while improving model performance.

    \subsection{Comparison with Domain Generalization and Adapation Methods} \label{sec:comp_with_DG_and_DA}
        \input{table/cross_dataset_3dhp_sota}
        We further compared the proposed method with various methods in DG and DA literature \footnote{The results of other methods are taken from DAF-DG  \cite{peng2024dual} and PoSynDA \cite{liu2023posynda}.}. 
        In the cross-dataset test on 3DHP (\cref{tab:cross_dataset_3dhp_sota}), the domain adaptation method PoSynDA achieved the best performance in terms of MPJPE. However, our proposed method demonstrated a comparable MPJPE error (58.2 mm vs. 59.5 mm) while achieving the best AUC performance through the combination of the canonical domain and the powerful DSTformer network. This outcome is particularly noteworthy as it was achieved without employing additional data augmentation techniques to enhance pose diversity. This suggests considerable potential for further performance improvements through the incorporation of data augmentation or adaptation techniques. In future work, we anticipate that synergies between the proposed canonicalization and data augmentation techniques could lead to more data-efficient augmentation strategies.

%% file: fig/fig6.tex
\begin{figure*}[t]
    \centering
    \includegraphics[width=0.8\linewidth]{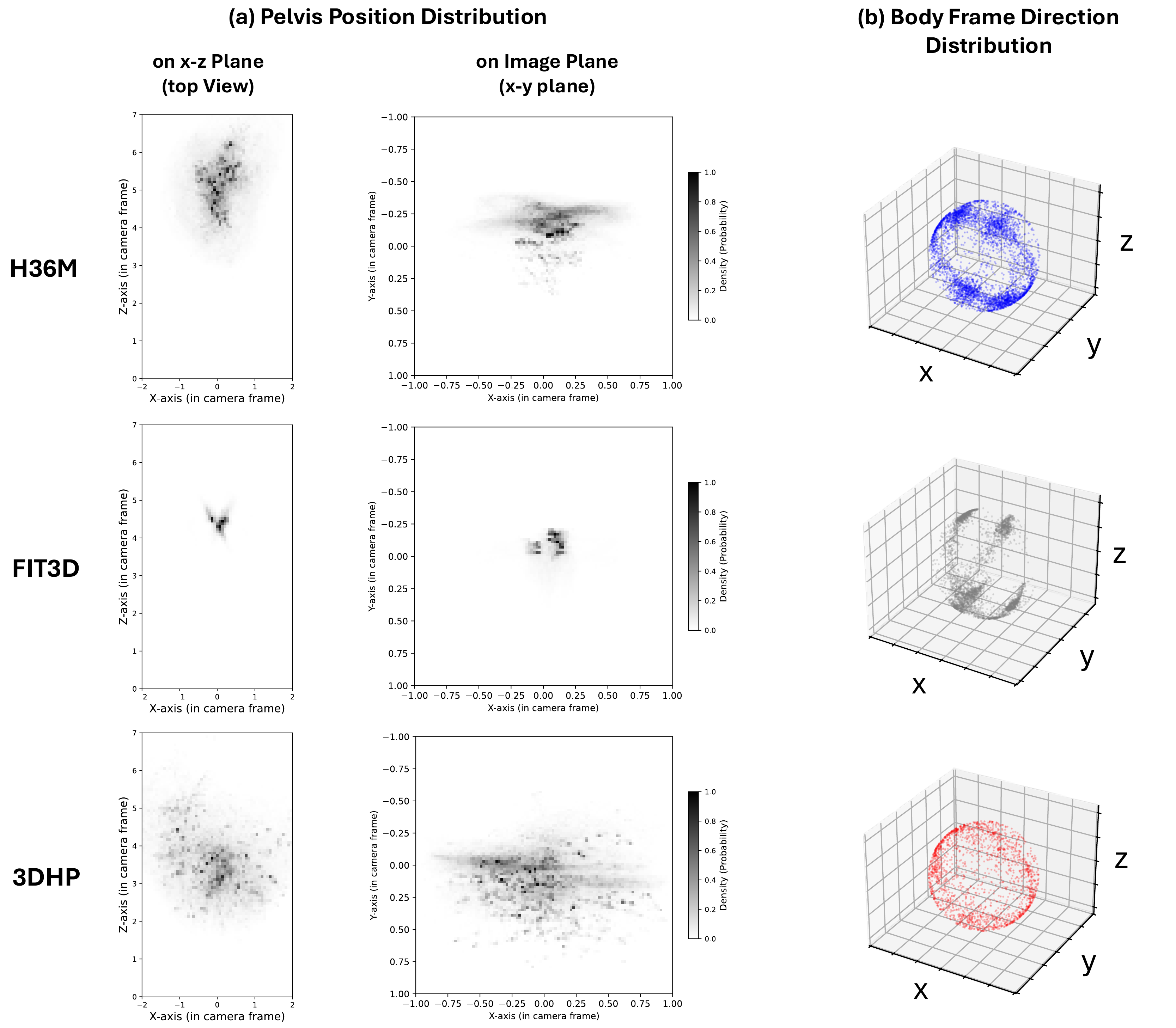}
    \caption{Camera-relative Pose (Position and Orientation) Distribution of Each Dataset: (a) Distribution of Pelvis Position in x-y plane and in image plane (Camera-relative Position), (b) Distribution of Unit Vector of Body-fixed Frame Direction (Camera-relative Orientation); The direction vector of body-fixed frame is calculated as the cross product of the vector from right hip to left hip joint and the vector from pelvis to torso joint.}
    \label{fig:fig6}
\end{figure*}

%% file: table/ablation_study.tex
\begin{table*}[]
\centering
\caption{Experiment Results for Ablation Study.}
\resizebox{\textwidth}{!}{%
\begin{tabular}{|c|c|c|c|c|c|c|c|c|}
\hline
\multirow{2}{*}{Train set} &
  \multirow{2}{*}{Test set} &
  \multirow{2}{*}{Method} &
  \multirow{2}{*}{\begin{tabular}[c]{@{}c@{}}2D \\ Canonical\end{tabular}} &
  \multirow{2}{*}{\begin{tabular}[c]{@{}c@{}}2D-3D Pose\\ Consistency\end{tabular}} &
  \multirow{2}{*}{\begin{tabular}[c]{@{}c@{}}MPJPE \\ (mm) $\downarrow$\end{tabular}} &
  \multirow{2}{*}{\begin{tabular}[c]{@{}c@{}}Error \\ reduction rate $\uparrow$\end{tabular}} &
  \multirow{2}{*}{\begin{tabular}[c]{@{}c@{}}P-MPJPE \\ (mm) $\downarrow$\end{tabular}} &
  \multirow{2}{*}{\begin{tabular}[c]{@{}c@{}}Error \\ reduction rate $\uparrow$\end{tabular}} \\
 &
   &
   &
   &
   &
   &
   &
   &
   \\ \hline
\multirow{6}{*}{\begin{tabular}[c]{@{}c@{}}H36M\\ S1\end{tabular}} &
  \multirow{3}{*}{\begin{tabular}[c]{@{}c@{}}FIT3D\\ ALL\end{tabular}} &
  Conventional &
  - &
  - &
  158.49 &
  - &
  94.57 &
  - \\
 &
   &
  Conventional + IC &
  o &
  - &
  130.05 &
  17.94\% &
  77.47 &
  18.08\% \\
 &
   &
  Canonical (Ours) &
  o &
  o &
  \textbf{110.21} &
  \textbf{30.46\%} &
  \textbf{71.20} &
  \textbf{24.71\%} \\ \cline{2-9} 
 &
  \multirow{3}{*}{\begin{tabular}[c]{@{}c@{}}3DHP\\ TEST SET\end{tabular}} &
  Conventional &
  - &
  - &
  87.75 &
  - &
  63.85 &
  - \\
 &
   &
  Conventional + IC &
  o &
  - &
  95.10 &
  -8.38\% &
  58.89 &
  7.77\% \\
 &
   &
  Canonical (Ours) &
  o &
  o &
  \textbf{70.46} &
  \textbf{19.70\%} &
  \textbf{57.53} &
  \textbf{9.90\%} \\ \hline
\multirow{6}{*}{\begin{tabular}[c]{@{}c@{}}H36M\\ S1, 5, 6, 7, 8\end{tabular}} &
  \multirow{3}{*}{\begin{tabular}[c]{@{}c@{}}FIT3D\\ ALL\end{tabular}} &
  Conventional &
  - &
  - &
  159.14 &
  - &
  101.75 &
  - \\
 &
   &
  Conventional + IC &
  o &
  - &
  92.68 &
  41.76\% &
  \textbf{67.44} &
  \textbf{33.72\%} \\
 &
   &
  Canonical (Ours) &
  o &
  o &
  \textbf{87.64} &
  \textbf{44.93\%} &
  70.87 &
  30.35\% \\ \cline{2-9} 
 &
  \multirow{3}{*}{\begin{tabular}[c]{@{}c@{}}3DHP\\ TEST SET\end{tabular}} &
  Conventional &
  - &
  - &
  64.87 &
  - &
  50.60 &
  - \\
 &
   &
  Conventional + IC &
  o &
  - &
  86.90 &
  -33.96\% &
  49.11 &
  2.94\% \\
 &
   &
  Canonical (Ours) &
  o &
  o &
  \textbf{59.53} &
  \textbf{8.23\%} &
  \textbf{47.79} &
  \textbf{5.55\%} \\ \hline
\end{tabular}%
}
\label{tab:ablation_study}
\end{table*}

%% file: table/additional_network.tex
\begin{table}[]
\centering
\caption{Experiment Results for Additional Lifting Network.}
\resizebox{\columnwidth}{!}{%
\begin{tabular}{|c|c|c|c|}
\hline
\multirow{2}{*}{Train set} &
  \multirow{2}{*}{Test set} &
  \multirow{2}{*}{Method} &
  \multirow{2}{*}{MPJPE (mm) $\downarrow$} \\
 &  &                       &                \\ \hline
\multirow{20}{*}{\begin{tabular}[c]{@{}c@{}}H36M\\ S1, 5, 6, 7, 8\end{tabular}} &
  \multirow{8}{*}{\begin{tabular}[c]{@{}c@{}}H36M\\ S9, 11\end{tabular}} &
  VideoPose + Conventional  &  43.11 \\
 &  & VideoPose + Canonical (Ours) & \textbf{42.39} \\ \cline{3-4} 
 &  & SB + Conventional     & \textbf{41.68} \\
 &  & SB + Canonical (Ours)        & 42.16          \\ \cline{3-4} 
 &  & SemGCN + Conventional & 49.75          \\
 &  & SemGCN + Canonical (Ours)    & \textbf{43.21} \\ \cline{3-4} 
 &  & ST-GCN + Conventional & 40.80          \\
 &  & ST-GCN + Canonical (Ours)    & \textbf{38.45} \\ \cline{2-4} 
 &
  \multirow{12}{*}{\begin{tabular}[c]{@{}c@{}}3DHP\\ TEST SET\\ (univ)\end{tabular}} &
  VideoPose + Conventional &
  82.93 \\
 &  & VideoPose + PoseAug   & 73.0           \\
 &  & VideoPose + Canonical (Ours) & \textbf{72.73} \\ \cline{3-4} 
 &  & SB + Conventional     & 81.58          \\
 &  & SB + PoseAug          & 76.2           \\
 &  & SB + Canonical (Ours)        & \textbf{74.83} \\ \cline{3-4} 
 &  & SemGCN + Conventional & 100.90         \\
 &  & SemGCN + PoseAug      & 86.1           \\
 &  & SemGCN + Canonical (Ours)    & \textbf{79.23} \\ \cline{3-4} 
 &  & ST-GCN + Conventional & 82.19          \\
 &  & ST-GCN + PoseAug      & \textbf{74.9}  \\
 &  & ST-GCN + Canonical (Ours)    & 76.45          \\ \hline
\end{tabular}%
}
\label{tab:additional_network}
\end{table}

%% file: table/cross_dataset_3dhp_sota.tex
\begin{table}[t]
\caption{Experiment Results for Cross-dataset on 3DHP Testset - DG: Domain Generalization, DA: Domain Adaptâtion, CA: Canonicalization.}

\centering
\label{tab:cross_dataset_3dhp_sota}
\resizebox{\columnwidth}{!}{%
\begin{tabular}{|c|c|c|c|c|c|}
\hline
Method                                  & Venue    & Type   & PCK $\uparrow$ & AUC $\uparrow$ & MPJPE $\downarrow$   \\ \hline
PoseAug \cite{gong2021poseaug}          & TPAMI'23 & DG     & 88.6      & 57.3      & 73                \\ 
DH-AUG \cite{huang2022dh}               & ECCV'22  & DG     & 89.5      & 57.9      & 71.2              \\ 
PoseGU \cite{guan2023posegu}            & CVIU'23  & DG     & 86.3      & 57.2      & 75                \\ 
CEE-Net \cite{li2023cee}                & AAAI'23  & DG     & 89.9      & 58.2      & 69.7              \\ 
DAF-DG \cite{peng2024dual}              & CVPR'24  & DG     & 92.9      & 60.7      & 63.1              \\ 
\hline
AdaptPose \cite{gholami2022adaptpose}   & CVPR'22  & DA     & 88.4      & 54.2      & 77.2              \\ 
PoseDA \cite{chai2023global}            & ICCV'23  & DA     & 92.1      & 62.5      & 61.3              \\ 
PoSynDA \cite{liu2023posynda}           & ACM'23   & DA     & \textbf{93.5}      & 59.6     & \textbf{58.2}     \\ 
\hline
Ours                                    & -        & CA     & 90.4       & \textbf{62.9}          & 59.5              \\
\hline
\end{tabular}}
\end{table}

%% file: sec/6_conclusion.tex
\section{Conclusion}
In this work, we addressed the challenge of domain gaps in 3D Human Pose Estimation (HPE) by introducing a novel canonical domain approach that unifies the source and target domains into a canonical domain, alleviating the need for additional fine-tuning in the target domain. The canonicalization process generates consistent 2D-3D pose mappings, ensuring 2D-3D pose consistency and simplifying pose patterns to enable more efficient training of lifting networks. With this approach, the lifting network is trained in the canonical domain, while input 2D poses in the target domain are canonicalized before inference using perspective projection properties and known camera intrinsics—bypassing the need for ground truth 3D data. We evaluated the proposed method using various lifting networks and publicly available datasets (e.g., Human3.6M, Fit3D, MPI-INF-3DHP), demonstrating substantial improvements in generalization capability across cross-dataset scenarios. These results confirm the effectiveness of the canonical domain approach in addressing domain gaps, offering a promising solution for improving generalization in 3D HPE.

In future work, we aim to combine the proposed canonicalization approach with advanced data augmentation techniques to further enhance generalization performance. Additionally, we plan to extend the canonicalization framework by incorporating factors such as scale and camera parameters to broaden its effectiveness.

%% file: sec/7_acknowledgement.tex
\section*{Acknowledgment}
This research was financially supported by the Ministry of Small and Medium-sized Enterprises(SMEs) and Startups(MSS), Korea, under the “Regional Specialized Industry Development Plus Program(R\&D, S3365742)” supervised by the Korea Technology and Information Promotion Agency (TIPA).